\documentclass[10pt,twocolumn,letterpaper]{article}

\usepackage{cvpr}
\usepackage{times}
\usepackage{epsfig}
\usepackage{graphicx}
\usepackage{amsmath}
\usepackage{amssymb}
\usepackage{multirow}

\usepackage[breaklinks=true,bookmarks=false]{hyperref}

\cvprfinalcopy 

\def\cvprPaperID{10550} 

\begin{document}

\title{Multi-view Human Pose and Shape Estimation Using Learnable Volumetric Aggregation}

\author{Soyong Shin \hspace{10mm} Eni Halilaj\\
Carnegie Mellon University\\
Pittsburgh, PA, USA\\
{\tt\small \{soyongs, ehalilaj\}@andrew.cmu.edu}
}

\maketitle

\begin{abstract}
   Human pose and shape estimation from RGB images is a highly sought after alternative to marker-based motion capture, which is laborious, requires expensive equipment, and constrains capture to laboratory environments. Monocular vision-based algorithms, however, still suffer from rotational ambiguities and are not ready for translation in healthcare applications, where high accuracy is paramount. While fusion of data from multiple viewpoints could overcome these challenges, current algorithms require further improvement to obtain clinically acceptable accuracies. In this paper, we propose a learnable volumetric aggregation approach to reconstruct 3D human body pose and shape from calibrated multi-view images. We use a parametric representation of the human body, which makes our approach directly applicable to medical applications. Compared to previous approaches, our framework shows higher accuracy and greater promise for real-time prediction, given its cost efficiency. 
\end{abstract}

\section{Introduction}

Accurate 3D human pose estimation could revolutionize the study and treatment of mobility-limiting medical conditions. In the field of biomechanics, accurate assessment of 3D pose is key in the early diagnoses of neuromusculoskeletal diseases and monitoring of rehabilitation. Conventionally, marker-based motion tracking systems or inertial measurement units (IMU) have been the dominant methods for human motion analysis. Although highly accurate, marker-based techniques have limitations, including intrusiveness, required expertise for data collection and analysis, and a need for expensive equipment, which make the approach not scalable across clinics. IMU-based approaches are portable, but relevant pose-estimation algorithms are susceptible to heading drift and a need for careful sensor-to-body calibration. Computer vision could address some of these limitations, but current algorithms lack the accuracy that is required for clinical applications.

\begin{figure}[t]
\begin{center}
\includegraphics[width=1.0\linewidth]{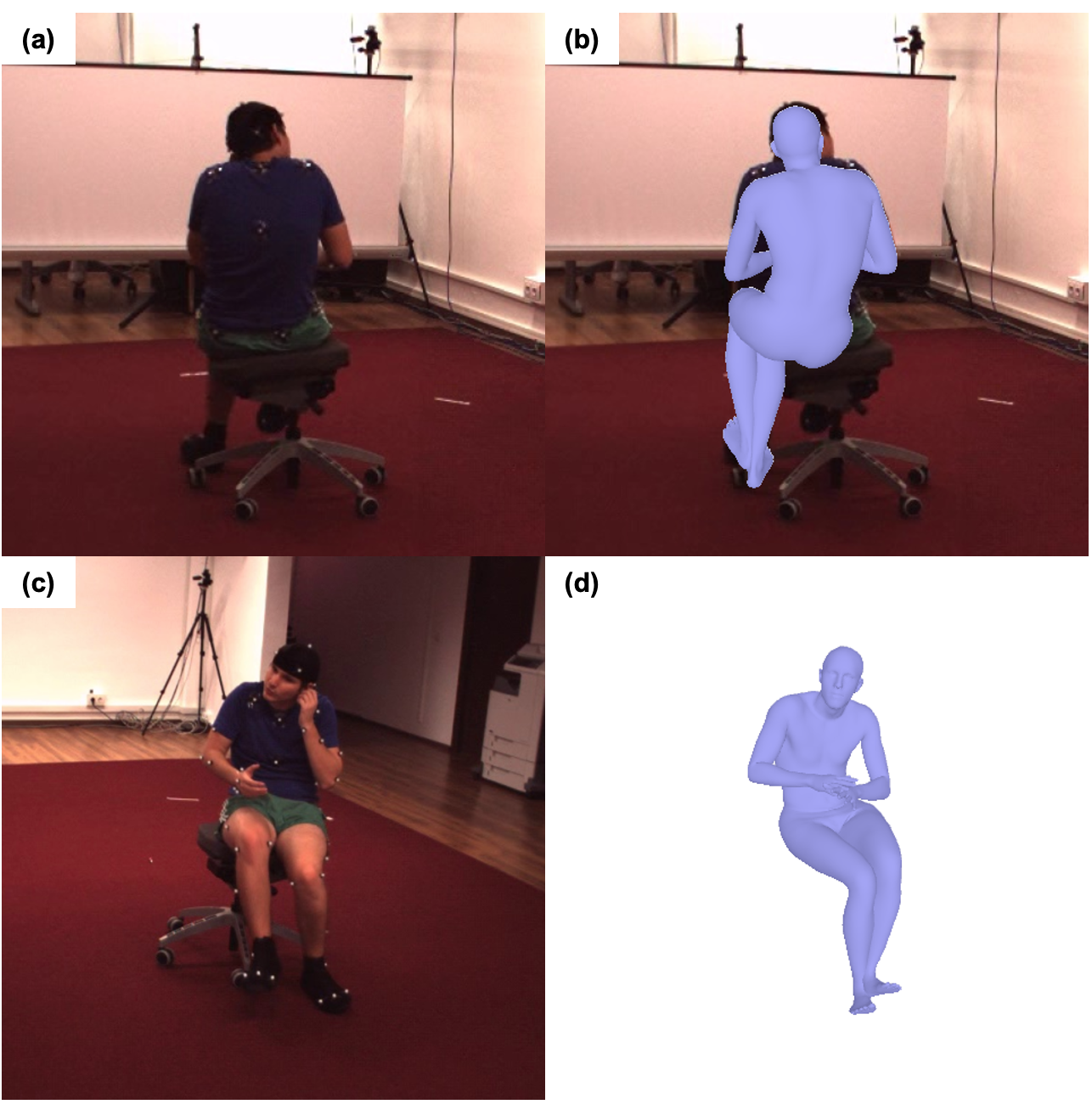}
\end{center}
   \caption{Erroneous predictions due to self-occlusion when using single-view pose estimation approaches. Predictions from single-view algorithms fit well on the input images, as shown in the example image (a) and predicted human body from this view (b), but may contain errors if there are occlusions, as shown in an image from another view (c) and the prediction from the second view (d), where it is clear the left hand position is not predicted correctly.}
\label{fig:long}
\label{fig:occlusion}
\end{figure}

Attempts at vision-based motion tracking are evolving from skeletal models based on joint locations \cite{insafutdinov2016deepercut, deepcut16cvpr, openpose, Moon_2018_CVPR} to statistical models that capture both body shape and 3D pose simultaneously \cite{kanazawaHMR18, kolotouros2019spin, Moon_2020}. Although skeletal-based approaches have improved in recent years, allowing both 2D and 3D joint center detection, they are not directly applicable in many medical applications. One of their major limitations is inability to capture 3D body pose, which bears significance in terms of disease progression or recovery. Deformable parametric body models are therefore more appropriate for biomechanics applications. Estimating human pose and shape from a single image using parametric body models has been a large area of focus in computer vision in recent years. The problem remains challenging and highly unconstrained, however, due to occlusions and lack of 3D information in 2D images (Figure \ref{fig:occlusion}). Optimization-based multi-view approaches have been explored as a potential solution, but these methods still do not meet the accuracy and computational efficiency requirements of clinical applications.

Inspired by a recently published multi-view aggregation algorithm that generates 3D keypoints using learnable triangulation \cite{iskakov2019learnable}, we propose a model-based volumetric aggregation method for 3D human pose and shape estimation. Our approach aggregates visual information from calibrated cameras in 3D global coordinate space using a back-project operation. It then estimates a single pose and shape of the human body from aggregated information using a 3D regression network. Unlike the previous method that depends on voxel-wise representations, our approach uses a human kinematics embedding that enables reconstruction of the human body from sparse information. This way, our approach does not require a large number of network parameters and can aggregate multiple images in a cost-efficient manner, which makes it convenient real-time pose estimation.

We evaluated our model on two multi-view image datasets, Human3.6M \cite{IonescuSminchisescu11} and MPI-INF-3DHP \cite{mono-3dhp2017}, and compared it with current state-of-the-art approaches. Our model outperforms existing single- and multi-view methods in 3D human pose and shape estimation. Our contributions are as follows:
\begin{itemize}
    \item [$\bullet$] {\bf Novelty:} We propose a 3D human pose and shape estimation model from multi-view RGB images using volumetric aggregation. Our model predicts body shape and joint angles, which are relevant for medical applications.
    \item [$\bullet$] {\bf Speed:} Our framework uses rich information from a parametric human body model, allowing aggregation in a computationally efficient way compared to the recently suggested voxel-wise 3D keypoint estimation.
    \item [$\bullet$] {\bf Accuracy:} Our aggregation method shows state-of-the-art performance in reconstructing 3D human body from multi-view images. Furthermore, the learnable aggregation method could enable utilization of the intermediate aggregated features for even better performance in the future.
\end{itemize}

\section{Related work}
In this section, we discuss related methods for predicting 3D human pose and shape from RGB images. We categorize these methods into two parts: 1) multi-view aggregation methods for 3D human pose estimation and 2) model-based approaches for reconstructing 3D body shape and pose.

\subsection{Pose estimation from multi-view images}
Multi-view approaches for human pose estimation have been generally used to generate ground truth 3D pose in the development of less accurate single-view algorithms. In the early stages, 3D pose was estimated by a simple triangulation of 2D pose from each view, generated with pretrained detectors \cite{openpose, insafutdinov2016deepercut, wei2016cpm}. The work of Kadkhodamohammadi \etal \cite{Kadkhodamohammadi_2020} reconstructs human pose in global 3D coordinates using a fully connected network, after concatenating 2D keypoints from all views into a single batch. To aggregate information from multiple views more efficiently, later studies then proposed deep learning architectures based on geometric information \cite{Qiu_2019_ICCV} \cite{Remelli_2020}.

\begin{figure*}[t]
\begin{center}
\includegraphics[width=0.9\linewidth]{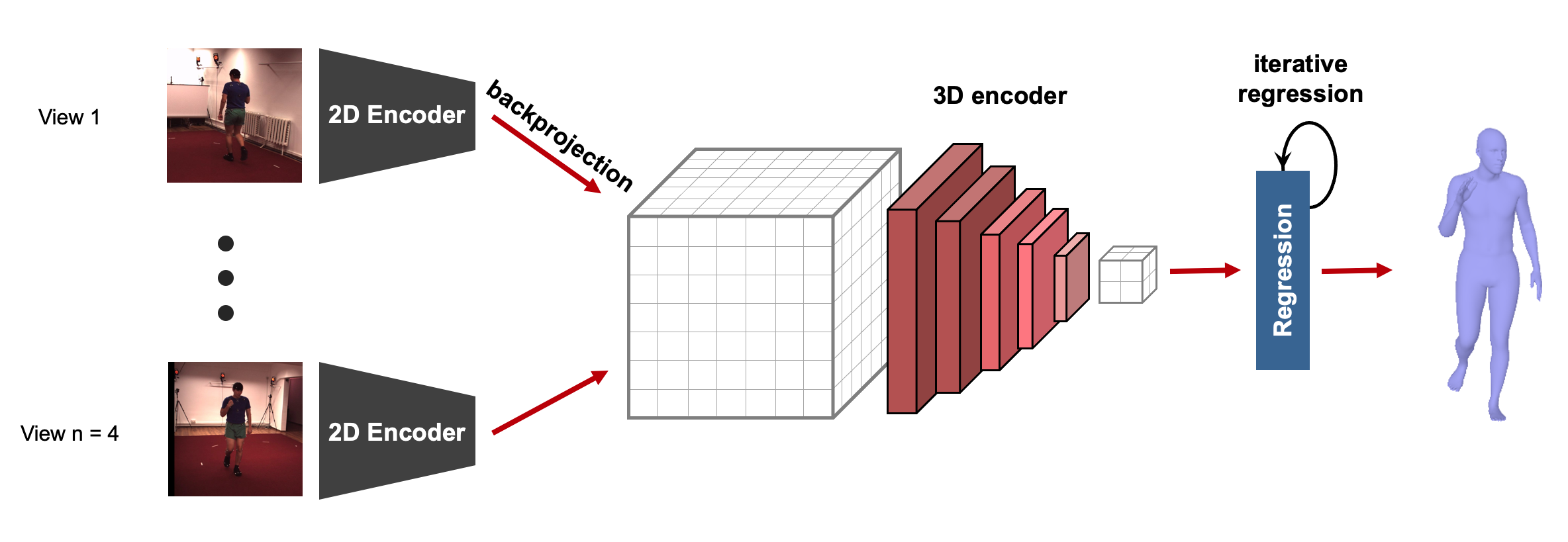}
\end{center}
  \caption{Overall pipeline of the approach. Figures from all views are encoded via a 2D backbone network, and then extracted feature maps are back-projected to a 3D global coordinate and aggregated into a single volume. SMPL parameters are the estimated through a regression network to finally reconstruct the full body pose.}
\label{fig:pipeline}
\end{figure*}

One major approach for 3D keypoint detection is to aggregate information from multiple viewpoints into volumetric 3D space \cite{Joo_2019, Pavlakos_2017, iskakov2019learnable}. Joo \etal \cite{Joo_2019} back-project multi-view 2D joint heatmaps from pretrained detectors into a common 3D space and estimates the probability of 3D joint locations in a global coordinate system. Iskakov \etal \cite{iskakov2019learnable} use end-to-end learnable triangulation to aggregate images from multiple views. Instead of 2D heatmaps, they back-project the intermediate feature maps of all views into a 3D grid and predict per-voxel heatmaps of 3D keypoints with a volumetric convolutional regressor. In addition to state-of-the-art performance, this geometry-based 3D approach shows promise for future applications since the triangulation is learnable, enabling incorporation of motion or pose priors. However, volumetric regression methods require high computational power to generate dense intermediate feature maps and 3D grid volumes for their voxel-wise representation \cite{iskakov2019learnable, Varol_2018, Jackson_2019}. In this work, we fuse information from multiple images using volumetric aggregation, but do so in a cost-efficient manner by employing a 3D kinematics embedding \cite{SMPL:2015}. By estimating 3D kinematics, this approach also opens the door to the translation of vision-based approaches in biomechanics and rehabilitation.

\subsection{Pose estimation with parametric body models}
Parametric 3D human body models \cite{SMPL:2015, Joo_2018, Pavlakos_2019} enable reconstruction of 3D body shape and pose from 2D images. One such model, Skinned Multi-Person Linear (SMPL), which learned salient information about 3D human shape and pose from a large 3D body-scan dataset \cite{CEASAR}, is widely used to solve 3D positional ambiguity from single or multiple images. Early-stage model-based approaches \cite{Bogo_2016, MuVS_2017} use optimization algorithms to fit the 3D body model to 2D keypoints predicted from a pretrained detector \cite{insafutdinov2016deepercut, openpose}. Since the optimization fitting is highly dependent on an initialization step and not translatable to real-time applications, learning-based regression models have been recently proposed. Pavlakos \etal \cite{Pavlakos_2018_CVPR} use a two-stage approach, first estimating 2D keypoints and a silhouette from a convolutional encoder and then separately regressing pose and shape parameters from two intermediate representations, respectively. Kanazawa \etal \cite{kanazawaHMR18} used a direct regression method to estimate a full set of SMPL parameters directly from the image and penalize infeasible human body predictions using an adversarial discriminator. Kolotouros \etal \cite{kolotouros2019spin}, based on \cite{kanazawaHMR18}, use direct supervision on SMPL parameters by incorporating an optimization fitting loop to generate semi-ground-truth labels. They can weakly penalize infeasible pose and shape, since the objective function of the fitting loop includes prior terms. These approaches have ushered progress in human pose and shape estimation from single-view images, but accuracies are still not optimal, especially when occlusion is present.

To address this challenge, multi-view approaches have also been explored. Kanazawa \etal \cite{MuVS_2017} used optimization fitting to find 3D human body parameters that minimize the reprojection error of joint locations and 2D silhouettes from each view. Liang \etal \cite{liang2019shape} proposed a stage-by-stage and view-by-view regression method. Following a structure similar to \cite{kanazawaHMR18}, their iterative algorithm transfers information on camera calibration to the next stage, while pose and shape prediction is transferred along each view. This method was the first to solve the multi-view aggregation for 3D pose and shape estimation in an end-to-end manner, but rather than aggregating the information from different views, their method simply expands the iteration loop and estimates SMPL parameters from all views. This approach not only requires longer inference time, as it takes larger iteration steps, but it also shows relatively low performance compared to the current state-of-the-art approach \cite{kolotouros2019spin}. By comparison, our approach aggregates the multiple images using geometric information from calibrated cameras and shows greater accuracy and shorter inference time.

\section{Method}
Figure \ref{fig:pipeline} shows the overall flow of our approach for reconstructing 3D human body pose and shape in a calibrated multi-view setting. The following sub-sections will describe (\ref{SMPL body mode}) the parametric 3D human body model, SMPL, (\ref{Aggregation method}) the aggregation method, which uses volumetric back-projection, and (\ref{Regression network}) the regression network.

\subsection{SMPL body model} \label{SMPL body mode}
Instead of estimating the full set of the human body mesh vertices directly, we estimate the parameters of SMPL \cite{SMPL:2015}, a generative model learned from a large scale 3D human body scan data \cite{CEASAR}. SMPL is a differentiable function, $\mathcal{M}(\theta, \beta)$, which maps a triangulated body mesh $M \in R^{N\times 3}$ to given pose $\theta \in R^{J\times 3}$ and shape $\beta \in R^{10}$ parameters. Pose parameters are a set of 3D rotation vectors representing body segments relative to their parent segments and the global orientation of the body (i.e., root joint rotation), for a total of $J=24$ vectors. The shape parameter vector represents the 10 directions of greatest shape variability, retrieved from Principle Component Analysis. The reconstructed 3D human body mesh consist of $N=6890$ vertices that can be linearly regressed to the 3D keypoints location $X \in R^{J'\times 3}$ from a pretrained regressor $W$ by $X=WM$. We use the linear regressor provided by \cite{kanazawaHMR18}, which outputs $J'=49$ keypoints. Unlike previous model-based approaches, we do not estimate camera translation since our approach directly estimates the 3D human body in a global coordinate system using the given camera calibration.

\begin{figure}[t]
\begin{center}
\includegraphics[width=1.0\linewidth]{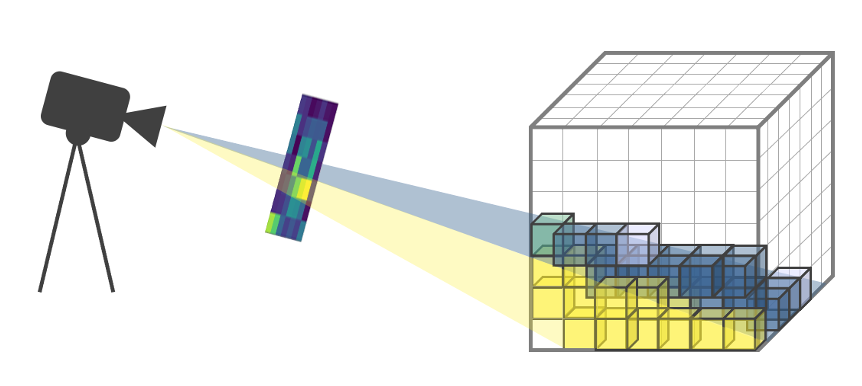}
\end{center}
  \caption{Back-project operation. Using the camera projection matrix, we back-project the feature maps into 3D space and fill the empty volume within the projection line.}
\label{fig:aggregation}
\end{figure}

\subsection{Aggregation method} \label{Aggregation method}
In order to fuse a subject’s pose and shape information captured from surrounding cameras, we aggregate 2D feature maps from each image into a 3D coordinate system through a back-projecting operation. The feature map $m_c$ of a 2D image ($I_c$) from camera view $c$ is extracted by a 2D convolutional encoder $e_{2D}$. We retrieve feature maps from a single 2D convolutional layer $g$ with a $1\times1$ kernel to downsize the channel before the aggregation. Down-sampled feature map $m_c^{in}$ is computed as
\[ m_c^{in} = g(e_{2D}(I_c)). \]
We then build a $K\times L\times L\times L$ cuboid $V^{coord}$ in a global coordinate system, where $K$ is channel size of $m_c^{in}$ and $L$ is the side length of the cuboid. The center point of the cuboid is set at the subject’s pelvis, which is obtained by algebraic triangulation of the 2D pelvis detection from each view. Following the back-project operation described in \cite{iskakov2019learnable}, we back-project $m^{in}_c$, the feature map from view $c$, into the established global coordinate volume and obtain the filled cube $V_c=B_c (m^{in}_c)$, where $B_c$ is the back-project operation using a projection matrix of camera $c$. 3D features from all the views are aggregated using a 3D softmax operation:
\[ V^{in} = \sum_{c} \left( \frac{\exp(V_c)}{\sum_{c} \exp(V_c) } \circ V_c \right ). \]
The framework of learning-based volumetric aggregation of intermediate feature maps has been previously introduced by \cite{iskakov2019learnable}, yielding state-of-the-art performance, but it requires a large number of voxels and high computational load since the final output of the network is a voxel-wise representation. We instead use a human kinematic embedding, as mentioned in section \ref{SMPL body mode}, aggregating relevant information with a much smaller number of parameters, which enables our method to make predictions in real-time.

\subsection{Regression network} \label{Regression network}
The final regressor is composed of 2 networks: a volumetric encoder, $e_{3D}$, and an iterative fully-connected regressor, $f^{reg}$. The aggregated volumetric map is encoded into a $2 \times 2 \times 2$ volume, $V^{out} = e_{3D}(V^{in})$. We then flatten the volume and obtain the final estimation, SMPL parameters $\Theta = \{ \theta, \beta \} $, using fully connected layers. Inspired by \cite{kanazawaHMR18}, we use a similar iterative structure that estimates final SMPL parameters $\Theta^{reg}=(\theta^{reg}, \beta^{reg})$ by iterative estimation: $\Theta_{i+1} = \Theta_i + \Delta \Theta_i$. However, our model does not estimate camera translation, since we assume that the camera calibration is given. We finally reconstruct the 3D human body mesh with $N=6890$ vertices, $M^{reg} = \mathcal{M}(\theta^{reg}, \beta^{reg})$. We train our model in an end-to-end manner using common 2D and 3D supervisions:
\[ L_{3D} = \big| \big| X_{3D}^{reg} - X_{3D}^{gt} \big| \big|, \]
\[ L_{2D} = \sum_{c} \big| \big| X_{c, 2D}^{reg} - X_{c, 2D}^{gt} \big| \big|, \]
where predicted 3D and 2D joint locations are computed by the pretrained joint regressor $\mathcal{W}$: 
\[ X_{3D}^{reg} = \mathcal{W}(M^{reg}), \hspace{3mm} X_{c, 2D}^{reg} = P_c X_{3D}^{reg},\]
where $P_c$ is the projection matrix of camera $c$. 

We also train our model with a loss term on SMPL parameters, which is beneficial for two reasons: (1) it is linear to our model prediction and (2) it can weakly penalize infeasible poses and shapes, since the labels are generated from a similar algorithm \cite{Bogo_2016} that uses a pose and shape prior function.
\[ L_{\Theta} = \big| \big| \Theta^{reg} - \Theta^{gt} \big| \big|\]
This loss is applied when the ground truth SMPL parameters $\Theta^{gt}$ are available. Otherwise, we use explicit prior terms, as given below:
\[ L_{prior} = L_{\theta} + L_{\beta}.\]
Here, we use a pose prior built on the CMU MoCap dataset \cite{CMU_MoCap}, which is parametrized based using the SMPL model with a previously proposed method, MoSh \cite{10.1145/2661229.2661273}. 
 The pose prior loss is approximately calculated using the minimum operator introduced by \cite{doi:10.1177/0278364913479413}:
\[L_{\theta} = \min_j (-log(cg_j \mathcal{N} (\theta ; \mu_{\theta,j}, \Sigma_{\theta, j}))). \]
We use $N=8$ Gaussians model given by \cite{Bogo_2016}.
For the shape prior $L_{\beta}$, we simply use $L2$ regularization:
\[L_{\beta} = \big| \big| \beta \big| \big|. \]
Again, we clarify that these explicit prior terms are applied only when the ground truth SMPL parameters are not available.

\begin{figure}[t]
\begin{center}
\includegraphics[width=0.9\linewidth]{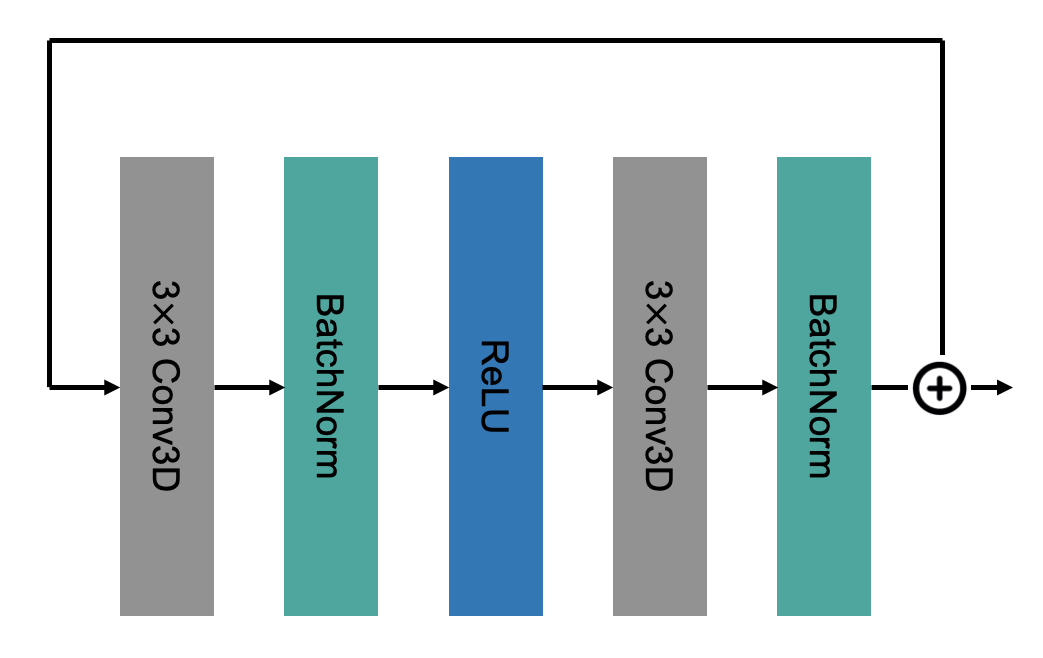}
\end{center}
  \caption{Architecture of the 3D residual block. The 3D encoder consists of several residual blocks and 3D pooling layers to encode the aggregated volume into a dense feature vector.}
\label{fig:residual_block}
\end{figure}

\subsection{Implementation details}
Datasets collected in a calibrated multi-view setting are required to train our learnable volumetric aggregation model, which uses camera extrinsics and intrinsics directly for the back-project operation. Rather than training the entire framework with these datasets, which were mostly captured indoors, we pretrained our 2D encoder using the same configuration proposed by \cite{kolotouros2019spin} on both outdoor and indoor datasets: LSP, LSP-Extension \cite{BMVC.24.12}, MS-COCO \cite{DBLP:journals/corr/LinMBHPRDZ14}, MPII \cite{6909866}, MPI-INF-3DHP, Human3.6M for 10 epochs to generalize the model. We then jointly trained the entire pipeline with multi-view datasets, Human3.6M and MPI-INF-3DHP, while setting the learning rate of the 2D encoder as 1/10 of the rest of the model. We confirmed that pretraining the backbone network significantly improves the performance of our model, even when tested on the indoor datasets. As described in \cite{iskakov2019learnable}, we also use the camera parameters of the Human3.6M dataset provided by Martinez \etal \cite{martinez_2017_3dbaseline} and crop input images using provided bounding box annotations.

We use ResNet50 \cite{DBLP:journals/corr/HeZRS15} for the 2D encoder ($e_{2D}$) architecture and aggregate the feature maps into the cuboid consisting $16\times16\times16$ voxels per each batch and channel after processing the feature maps with output channel $K=256$. The length of the volume in global coordinates is set as 2500 \textit{mm} to ensure that the projected information is thoroughly contained in the volume. The center of the volume is calculated using algebraic triangulation of the 2D pelvis location, detected via OpenPose \cite{openpose} in each camera view. We build 3D encoder $e_{3D}$ composed of several residual blocks and 3D pooling layers, where the architecture of each residual block is described in Figure \ref{fig:residual_block}. Also, final regression network $f^{reg}$ consists of 2 fully connected layers with 4-steps iterative regression method introduced in \cite{kanazawaHMR18}. We set the learning rate of pretrained 2D backbone to $10^{-5}$ and the rest of the model to $10^{-4}$ and used Adam solver \cite{kingma2014method}.

\begin{figure*}[p]
\begin{center}
\includegraphics[width=1.0\linewidth]{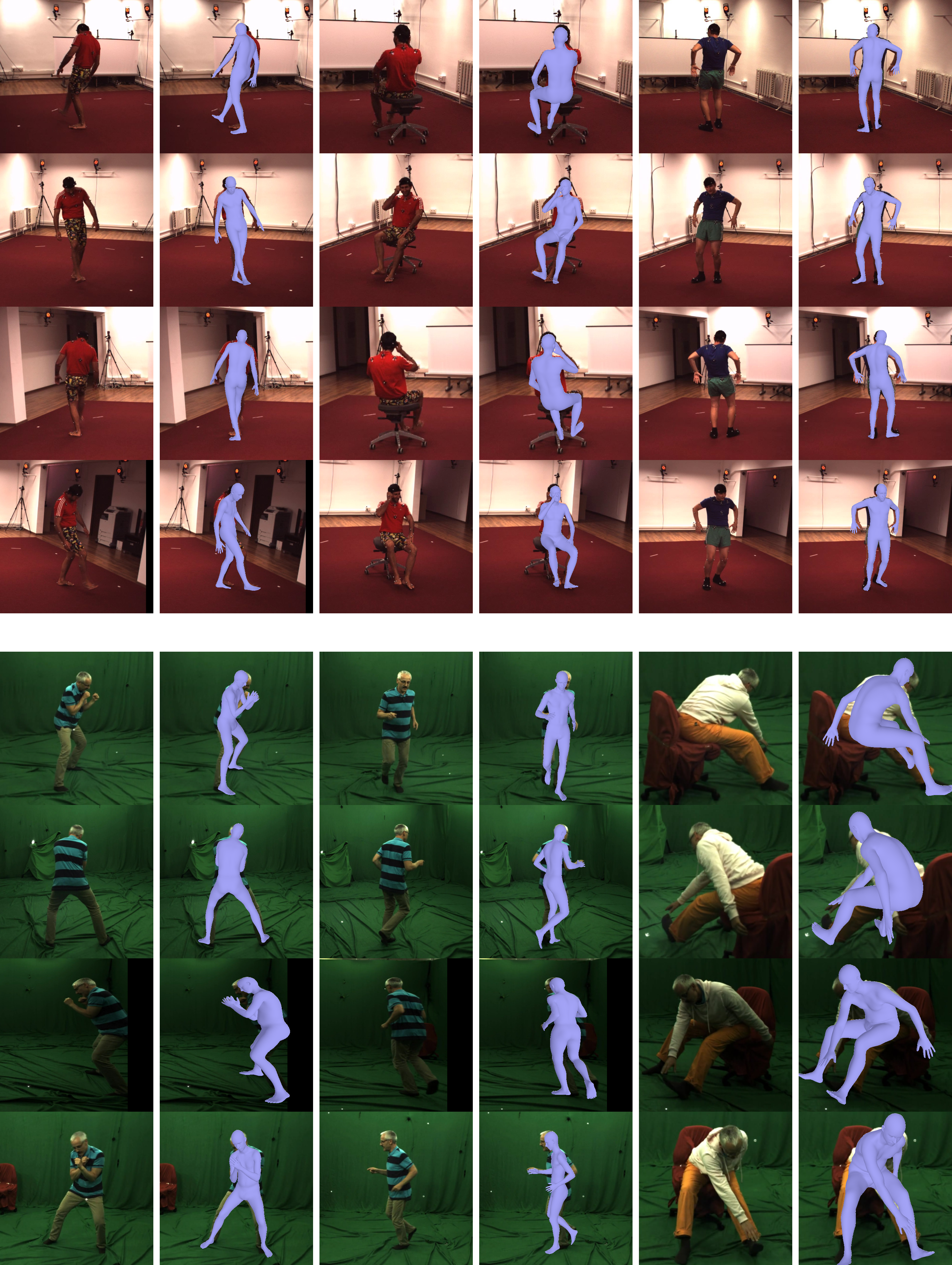}
\end{center}
\caption{Qualitative results of our approach on Human3.6M (rows 1-4) and MPI-INF-3DHP (rows 5-8). The model takes 4 multi-view images as input and predicts a single reconstructed human body overlaid to each camera view.}
\label{fig:sample_fig}
\end{figure*}

\section{Experiments}
We conducted experiments on two publicly available multi-view datasets, Human3.6M and MPI-INF-3DHP. For the first experiment, we trained our model on Human3.6M dataset for 20 epochs and for the second experiment, we fine-tuned on MPI-INF-3DHP for another 10 epochs.

\subsection{Datasets and evaluation metrics}
\textbf{Human3.6M.}
This is one of the largest benchmarks for indoor 3D human pose estimation. The dataset includes 3.6 million video frames from 4 different viewpoints, captured at 50Hz frames per seconds. The 3D ground truth annotation was collected using a marker-based motion capture system with 10 infrared cameras. \textit{Mean per joint position error} (MPJPE), the Euclidean distance between the prediction of our model and the given 3D joint labels after root joint alignment (in $mm$) is used for the evaluation. We also evaluated the performance after applying Procrustes analysis, by computing a similarity transform and aligning the prediction to the ground truth. We denote 3D error after the alignment as PA MPJPE and compare with prior methods as well. We generated labels of SMPL parameters using the 3D version of SMPLify and used it for direct supervision on SMPL parameters, as the 3D error of the label is much smaller than the error of our prediction. We trained our model using subjects S1, S5, S6, S7, S8, and tested on subjects S9 and S11.

\textbf{MPI-INF-3DHP.}
 MPI-INF-3DHP dataset is another multi-view dataset, mostly captured indoors using a green-screen studio and 14 cameras. Only a subset of 4 cameras, however, is used for both training and testing. Unlike Human3.6M, this dataset does not incorporate a marker-based motion capture system. As a result, the 3D label is relatively inaccurate. We evaluated our model's performance after applying Procrustes analysis using the same metric that was used for the Human3.6M dataset: MPJPE. We also use \textit{Percentage of Correct Keypoints} (PCK) and \textit{Area Under the Curve} (AUC) to compare against other multi-view approaches after applying Procrustes analysis. AUC is calculated with interval of 5 $mm$, following \cite{liang2019shape}. For this dataset, we did not achieve enough accuracy for SMPL parameter labels and therefore did not use direct supervision, but applied the explicit prior terms described at Section \ref{Regression network}. We trained our model on subjects S1 to S7 and tested on subject S8.

\subsection{Quantitative evaluation}

\begin{table}
\begin{center}
\resizebox{0.5\textwidth}{!}{
\begin{tabular}{|l|c|c|c|c|}
\hline
 & \multirow{2}*{MPJPE} & \multirow{2}*{PA MPJPE} & camera & feature \\
 & & & calib. & aggregation \\
\hline\hline
SPIN \cite{kolotouros2019spin} & - & 41.1 & No & No \\
MuVS \cite{MuVS_2017} & 58.22 & 47.09 & Yes & No \\
Liang \etal \cite{liang2019shape} & 79.85 & 45.13 & No & No \\
SPIN$^4$ & 57.5 & 35.4 & No & No \\
SPIN$^{4, cal}$ & 49.8 & 35.4 & Yes & No \\
Ours & \textbf{46.9} & \textbf{32.5} & Yes & Yes \\
\hline
\end{tabular}}
\end{center}
\caption{Comparison of different models on the Human3.6M dataset (in \textit{mm}), with each model trained on different datasets. The second column MPJPE is calculated after Procrustes analysis (PA). Our model shows better result before and after the PA alignment.}
\label{table:eval-h36m}
\end{table}

Here, we present our results and compare them with other multi-view approaches. For fair comparison with SPIN \cite{kolotouros2019spin}, which is a single-view approach that shows state-of-the-art performance in model-based human body reconstruction, we implemented a multi-view version. We independently predict SMPL parameters from each camera view and average pose and shape along the axis of 4 viewpoints. Since body orientation is dependent on camera pose, we do not average it. Instead, we take two different strategies to estimate body orientation. The first uses only the prediction from the frontal camera, assuming it includes more information than any other view. This approach does not require camera calibration, and we denote it as SPIN$^4$, given that the final reconstruction uses 4 input images. The second strategy is to transform body orientation predictions from all the cameras to a global coordinate system and average them. This coordinate transformation requires calibration. We thus denote it as SPIN$^{4, cal}$.

Although each model was trained on the different datasets, for performance comparison, we tested on the same datasets. In Table \ref{table:eval-h36m}, we present the performance of our model on the Human3.6M dataset and compare it against state-of-the-art regression methods \cite{kolotouros2019spin, liang2019shape} and an optimization algorithm \cite{MuVS_2017}. Our approach outperform the previous models both before and after Procrustes analysis. 

\begin{table}
\begin{center}
\resizebox{0.4\textwidth}{!}{
\begin{tabular}{|l|c|c|c|}
\hline
 & PCK & AUC & MPJPE \\
\hline\hline
Liang \etal \cite{liang2019shape} &  95 & 63 & 62 \\
Ours & \textbf{97.4} & \textbf{65.5} & \textbf{50.2} \\
\hline
\end{tabular}}
\end{center}
\caption{Comparison of models on the MPI-INF-3DHP dataset, where each model is trained on different datasets. Evaluation was calculated after aligning the prediction using Procrustes analysis. Higher PCK / AUC and lower MPJPE stands for better results. Our approach outperforms the previous multi-view method.}
\label{table:eval-mpi}
\end{table}

Similarly, we present our result on the MPI-INF-3DHP dataset in Table \ref{table:eval-mpi}. Here we do not compare against SPIN \cite{kolotouros2019spin}, since the publicly available version of SPIN has been trained on all multi-view data from MPI-INF-3DHP, including the test subject that we used to evaluate our mode. These comparisons indicate that our model outperforms prior model-based approaches with multiple images.

\begin{table}[h]
\begin{center}
\resizebox{0.5\textwidth}{!}{
\begin{tabular}{|l|c|c|c|}
\hline
 & number of & GPU memory & inference \\
& parameters & usage & time \\
\hline\hline
Liang \etal \cite{liang2019shape}&  27M & 1.30MB & 29ms \\
SPIN$^{4}$ & 27M & 1.32GB & 6ms \\
Iskekov \etal \cite{iskakov2019learnable} & 86M & 6.08GB & 91ms \\
Ours & 72M  & 1.98GB & 14ms \\
\hline
\end{tabular}}
\end{center}
\caption{Comparison of computational costs. We compare our model with the other aggregation methods in terms of the number of model parameters, model size, and inference time.}
\end{table}

We further evaluated our model's computational efficiency and compared against state-of-the-art approaches for both 3D keypoints detection and model-based 3D human pose reconstruction. Using GeForce RTX 2080Ti GPU, we measured an average of 100 iterations of inference time and GPU usage of each model with 4 input images with a shape of 224 $\times$ 224. Under the similar aggregation architecture, our model-based approach significantly reduces GPU memory usage and inference time compared to the previous learnable triangulation approach \cite{iskakov2019learnable}. Furthermore, our model has comparable computational load with other model-based approaches \cite{kolotouros2019spin, liang2019shape} in terms of memory usage. We even achieve faster inference than stage-by-stage and view-by-view estimation \cite{liang2019shape}. This result shows that the learnable volumetric aggregation for model-based 3D body reconstruction yields both high precision and real-time inference.

\begin{figure}[t]
\begin{center}
\includegraphics[width=1.0\linewidth]{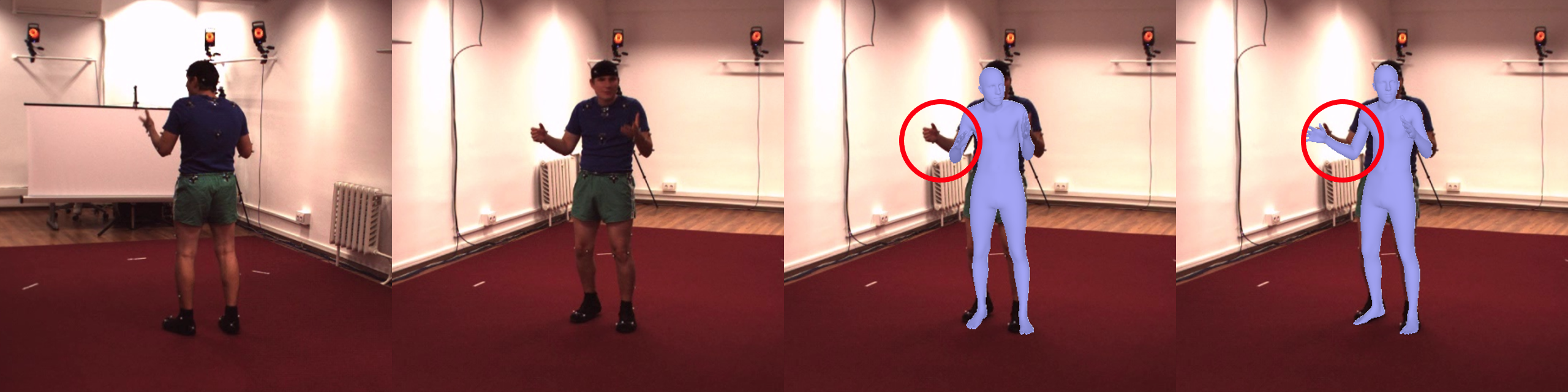}
\includegraphics[width=1.0\linewidth]{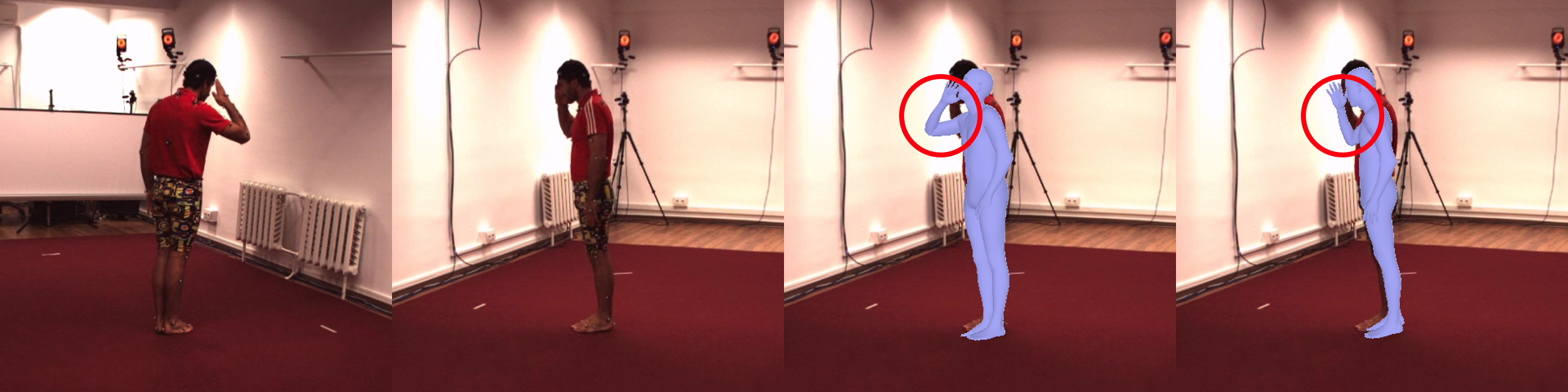}
\includegraphics[width=1.0\linewidth]{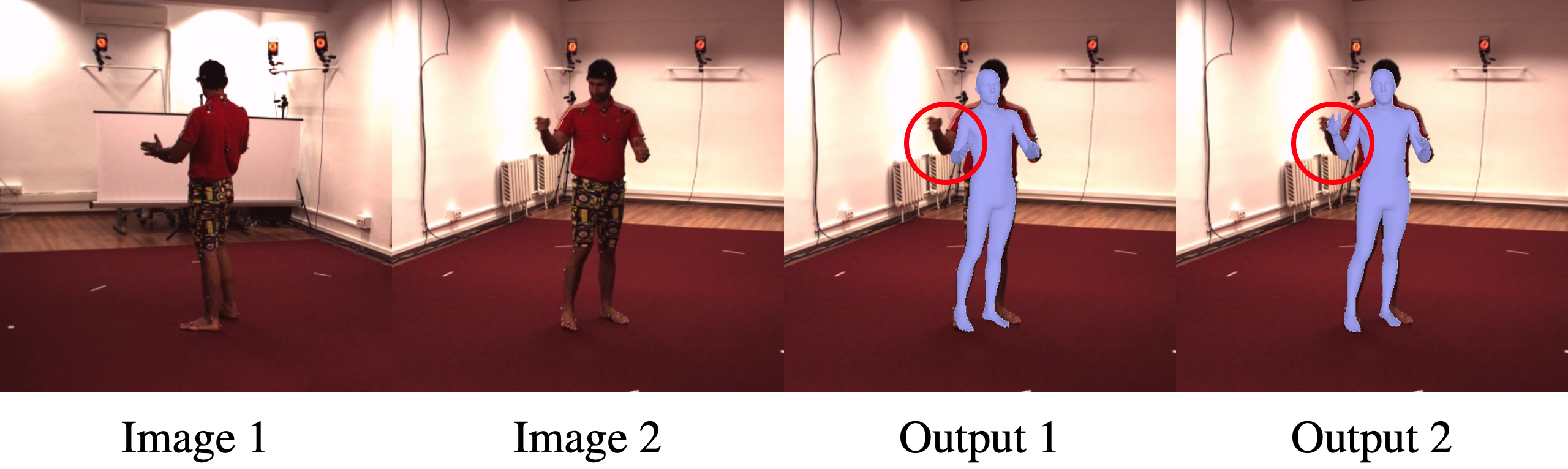}
\end{center}
\caption{
Qualitative comparison of our method with multi-view input images versus occluded single image. Image 1: input image for single-view model, Image 2: image from the other view, Output 1: prediction from single-view model, Output 2: prediction from multi-view model. Red circles show how multi-view model benefits from different views.}
\label{fig:comparison_with_single_view}
\end{figure}

We also report qualitative results of our approach on both datasets, Human3.6M and MPi-INF-3DHP in Figure \ref{fig:sample_fig}. The reconstructed 3D human body overlaid on original images from each view might seem less accurate compared to the reported figures of single-view approaches \cite{kanazawaHMR18, kolotouros2019spin}. However, this is natural since the primary goal of single-view approaches is to reconstruct a 3D human body that is the best fit with the given view. Our model, however, reconstructs a human body that aligns with all input images. Also, we report qualitative comparison of our method using multi-view images against only a single-view image with occlusion in Figure \ref{fig:comparison_with_single_view}.

\begin{figure}[t]
\begin{center}
\includegraphics[width=0.9\linewidth]{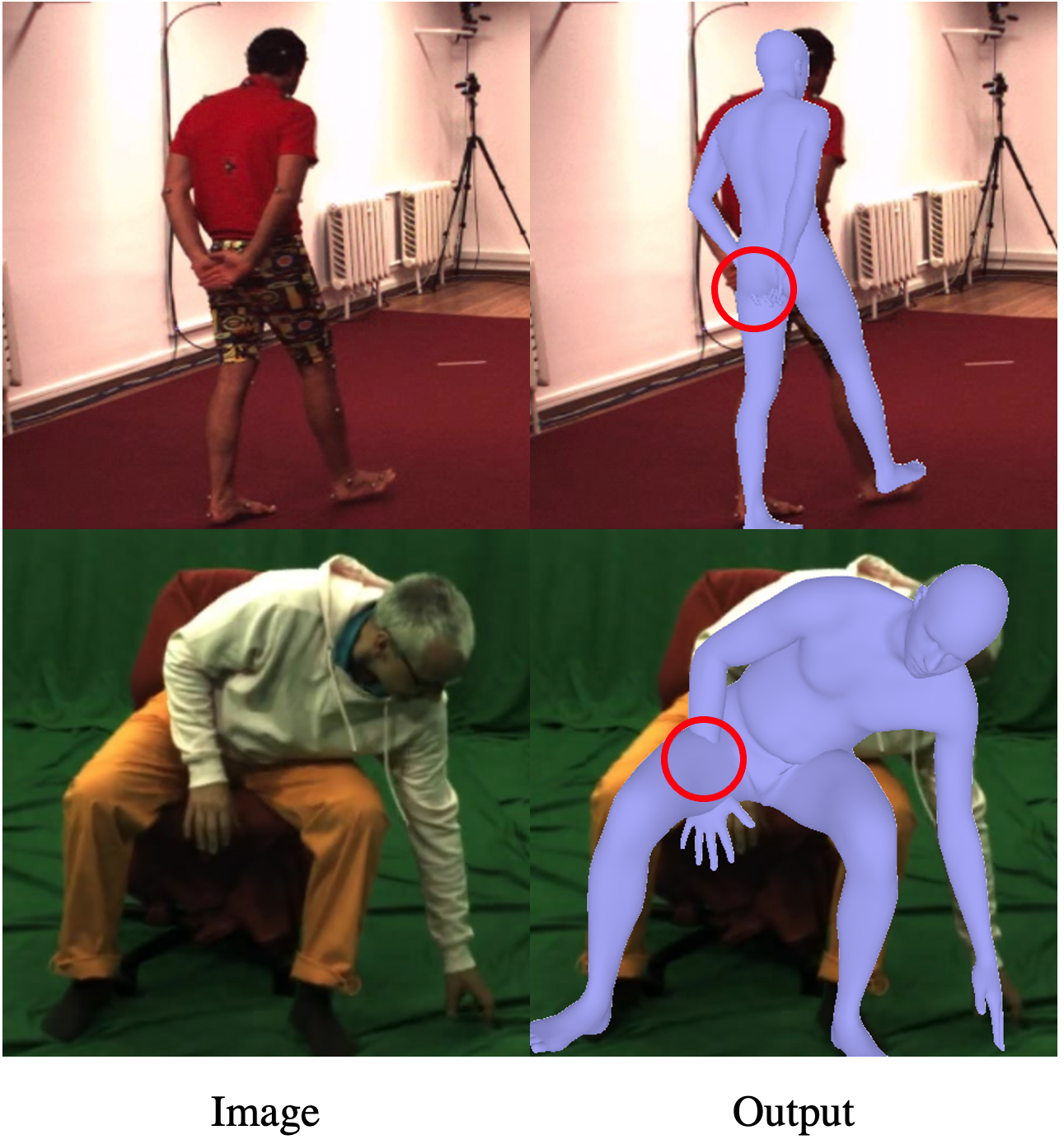}
\end{center}
\caption{Examples of erroneous predictions that have self-collision. Figures on right are the original pose of the subjects, while figures on left are the reconstructed body poses. Red circles show the regions of self-collision.}
\label{fig:sample_figure_with_collision}
\end{figure}

\section{Limitations}
Despite showing the state-of-the-art performance, our method still has limitations. One is self-collision. Similar to other model-based regression methods, we do not apply any penalty function for collision (or segment penetration), and therefor the prediction might suffer from self-collision. Figure \ref{fig:sample_figure_with_collision} illustrates examples of self-collision, where 3D errors are comparable to the average (less than 50 $mm$ before applying Procrustes analysis). These erroneous predictions usually happen when the two different body segments of the subjects are contacting each other. In this case, it is difficult to penalize these pose by only relying on prior distribution of human pose, since the set of predicted pose might be close enough to the ground truth. In this regard, additional penalty functions need to be introduced to prevent collisions. Another limitation, in the context of translating vision-based approaches to healthcare applications, is that we report joint center errors, but not errors in 3D kinematics (i.e., joint angles). Biomechanists and rehabilitation specialists are trained to interpret mobility limitations in terms of joint kinematics. In the future more emphasis may be placed on expanding the metrics used to evaluate and compare computer vision models, allowing the medical community to assess their transnational potential.

\section{Conclusion}
We propose a learnable aggregation method for 3D human body shape and pose estimation from multi-view RGB images. Our method improves previous approaches in three ways. First, it uses salient information from a parametric body model to improve computational efficiency compared to the priorly proposed learnable volumetric aggregation, which uses voxel-wise predictions. This makes the model ideal for real-time inference. Second, it achieves state-of-the-art performance in multi-view human pose and shape estimatio. Our method significantly outperforms prior methods on benchmark datasets. Last, the proposed feature-level aggregation approach is promising for future work in motion capture with multi-view cameras. Although previously proposed methods can still take multi-view input images and benefit from different views, to the our best knowledge, there is no approach that aggregates extracted features into a global space containing information from all views. Since current motion analysis networks, mostly based on a single-view video, estimate human motion using per-frame features from temporal regression networks \cite{Kanazawa_2019, Zhang_2019, Kocabas_2020}, our approach is more likely to be incorporated with video inference methods and advance video-based motion analysis in multi-view settings.

{\small
\bibliographystyle{ieee_fullname}
\bibliography{egbib}
}

\end{document}